\documentclass[11pt]{article}

\usepackage[final]{acl}
\usepackage{tabularx}

\usepackage{times}
\usepackage{latexsym}

\usepackage[T1]{fontenc}    
\usepackage[utf8]{inputenc} 

\usepackage{microtype}

\usepackage{inconsolata}

\usepackage{graphicx}
\usepackage{multirow}
\usepackage{array}  

\title{\Large{\textbf{Chain-of-Translation Prompting (CoTR): A Novel Prompting Technique for Low Resource Languages}}}

\author{Tejas Deshpande\textsuperscript{1, *}, Nidhi Kowtal\textsuperscript{1, *} \\
\textbf{Raviraj Joshi}\textsuperscript{2,3} \\
  \textsuperscript{1} Pune Institute of Computer Technology, Pune, Maharashtra India \\ 
  \textsuperscript{2} Indian Institute of Technology Madras, Chennai, Tamil Nadu India\\ 
  \textsuperscript{3} L3Cube Labs, Pune\\
  \texttt{\{tejasdeshpande1112, kowtalnidhi\}@gmail.com} \\
  \texttt{ravirajoshi@gmail.com}}

\begin{document}
\maketitle

\begin{abstract}
This paper introduces Chain of Translation Prompting (CoTR), a novel strategy designed to enhance the performance of language models in low-resource languages. CoTR restructures prompts to first translate the input context from a low-resource language into a higher-resource language, such as English. The specified task like generation, classification, or any other NLP function is then performed on the translated text, with the option to translate the output back to the original language if needed. All these steps are specified in a single prompt. We demonstrate the effectiveness of this method through a case study on the low-resource Indic language Marathi. The CoTR strategy is applied to various tasks, including sentiment analysis, hate speech classification, subject classification and text generation, and its efficacy is showcased by comparing it with regular prompting methods. Our results underscore the potential of translation-based prompting strategies to significantly improve multilingual LLM performance in low-resource languages, offering valuable insights for future research and applications. We specifically see the highest accuracy improvements with the hate speech detection task. The technique also has the potential to enhance the quality of synthetic data generation for underrepresented languages using LLMs.
\end{abstract}

\section{Introduction}

Natural Language Processing (NLP) has made significant progress in recent years, with models capable of understanding, creating, and translating human language across a wide range of tasks and languages. However since high-resource languages like English, Spanish, and Chinese have access to a wealth of annotated datasets and linguistic resources, most of this development has been focused on those languages. Low-resource languages, on the other hand, have a lot more difficulties since they lack large-scale and high-quality datasets \cite{b1}. Training effective NLP models are challenging due to this data scarcity, which frequently leads to subpar performance and poor generalization. Low-resource languages have distinct grammatical structures, linguistic diversity, and cultural quirks that make it more difficult to create accurate models and limit their use in practical contexts \cite{b2}. Multilingual LLMs have limitations on processing the prompts in low-resource languages \cite{b33}. This is because the amount of data used to train or fine-tune the model is very less. As a result, speakers of low-resource languages are frequently excluded from the benefits of advanced NLP technologies, highlighting the crucial need for novel techniques to close this gap.
\newline

\begin{figure}[]
  \centering
\includegraphics
[width=\linewidth]
{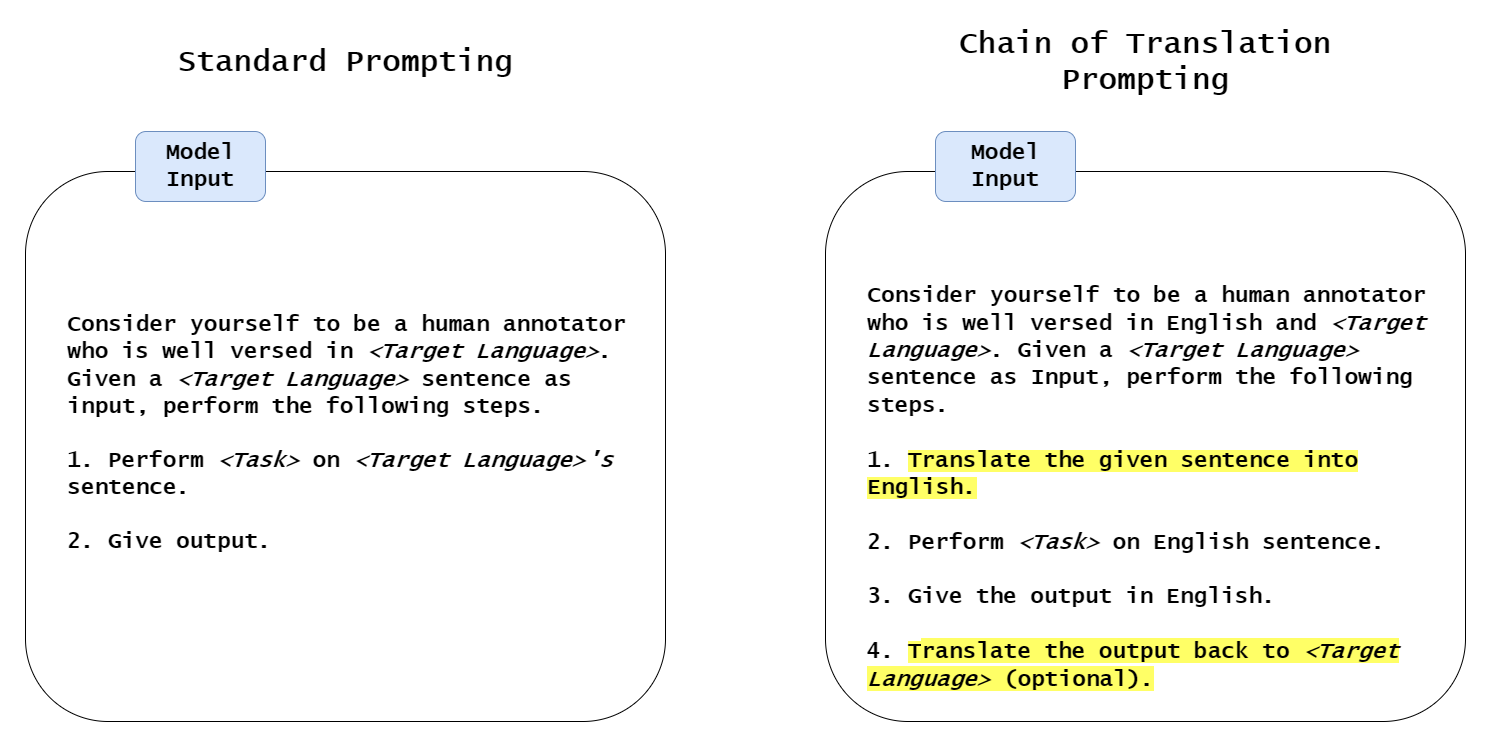}
\caption{A brief overview of the Chain of Translation Prompting (CoTR) for an annotation task. The technique modifies the input prompt to encapsulate the translation of the non-English input context to English, followed by performing the target task on the translated text.}
\label{fig:overview}
\end{figure}
However, Multilingual LLMs are good at translation tasks, as it is common practice to include parallel corpora during the pre-training stage \cite{b45}. We can leverage the ability of multilingual LLMs to improve responses for low-resource languages. In our study, we apply this approach to Marathi, an Indo-Aryan language spoken by about 83 million people, primarily in the Indian state of Maharashtra. Marathi is one of these low-resource languages \cite{b46,b48}. 
Despite its large speaker base, Marathi lacks digital resources, annotated corpora, and computational tools. The language's complex syntax challenges the development of precise NLP models, and limited Marathi-specific datasets and pre-trained models hinder the adoption of language technologies \cite{b3}. Therefore, new approaches are needed to enhance Marathi NLP performance and enable its speakers to benefit from AI advancements.

In this work, we investigate new prompting strategies to enhance Marathi language processing capabilities in models such as GPT-4o, GPT-4o Mini, Llama3-8B, Llama3-405B, and Gemma-9B. Our research introduces a novel strategy called "Chain of Translation Prompting (CoTR)", which we evaluate against direct Marathi prompting. We apply this method to sentiment analysis, hate speech classification, and subject categorization across three datasets: MahaSent \cite{b42,b42_2}, MahaHate \cite{b43}, and MahaNews-SHC \cite{b44,b44_2} respectively. Additionally, we assess its effectiveness in generating headlines using the CSEBUETNLP XLSum dataset. Our findings reveal that translating Marathi input into English and then performing classification or text generation using a single prompt yields superior results compared to directly processing the Marathi text with a standard prompt. This work significantly contributes to multilingual NLP by demonstrating the potential of translation-based prompting strategies, particularly with a single prompt, to enhance NLP performance in low-resource languages.

The main contributions of this work are as follows:
\begin{itemize}
    \item We introduce Chain of Translation Prompting (CoTR) as a method for performing input context translation during LLM response generation. Our results demonstrate that CoTR consistently outperforms standard prompting strategies across a variety of models and datasets.
    \item We benchmark various open and closed LLMs, including GPT-4o, GPT-4o mini, Llama 3.1 405B, Llama 3.1 8B, and Gemma 2 9B, on tasks such as Marathi Sentiment Analysis, Hate Speech Detection, News Categorization, and News Headline Generation. In terms of performance, closed LLMs consistently rank higher: GPT-4o > GPT-4o mini > Llama 3.1 405B > Gemma 2 9B > Llama 3.1 8B. We observe that CoTR is particularly beneficial for smaller models with higher error rates.
    \item The CoTR prompting strategy shows the most significant improvements in complex tasks like hate speech detection and sentiment analysis.
\end{itemize}

\begin{figure*}[htbp]
  \centering
  \includegraphics[width=0.8\textwidth]{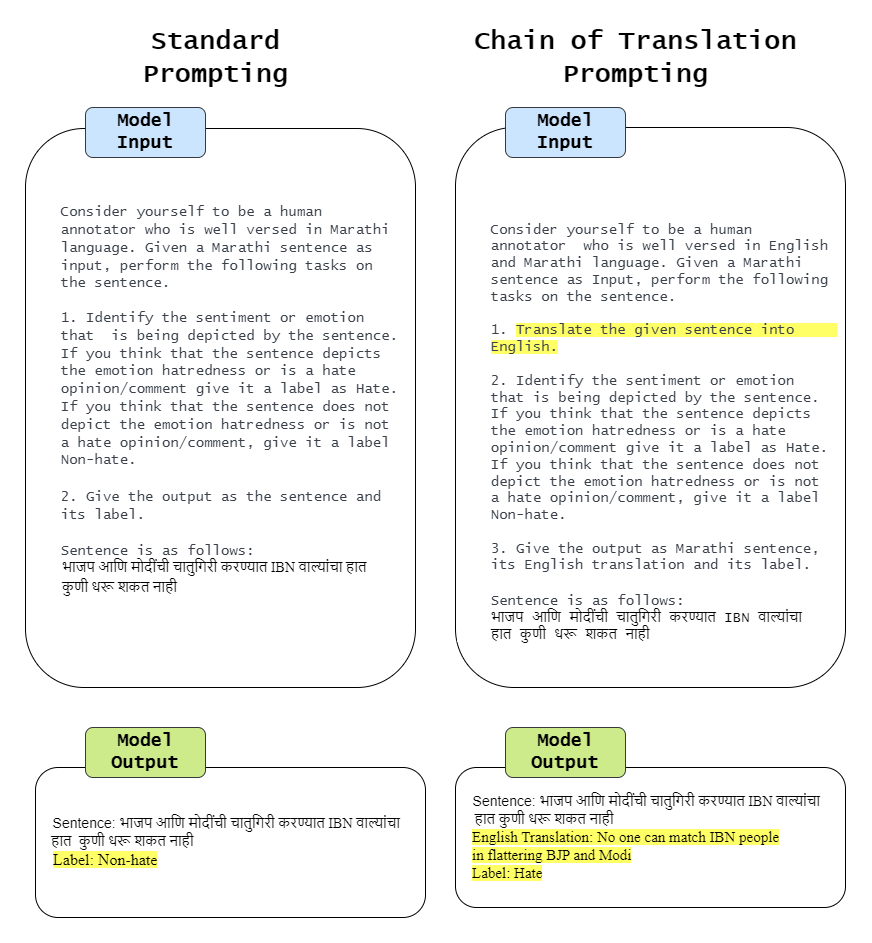}
  \caption{Prompt for Classification Task}
  \label{fig:ict_paper_fig}
\end{figure*}

\begin{figure*}[htbp]
  \centering
  \includegraphics[width=0.8\textwidth]{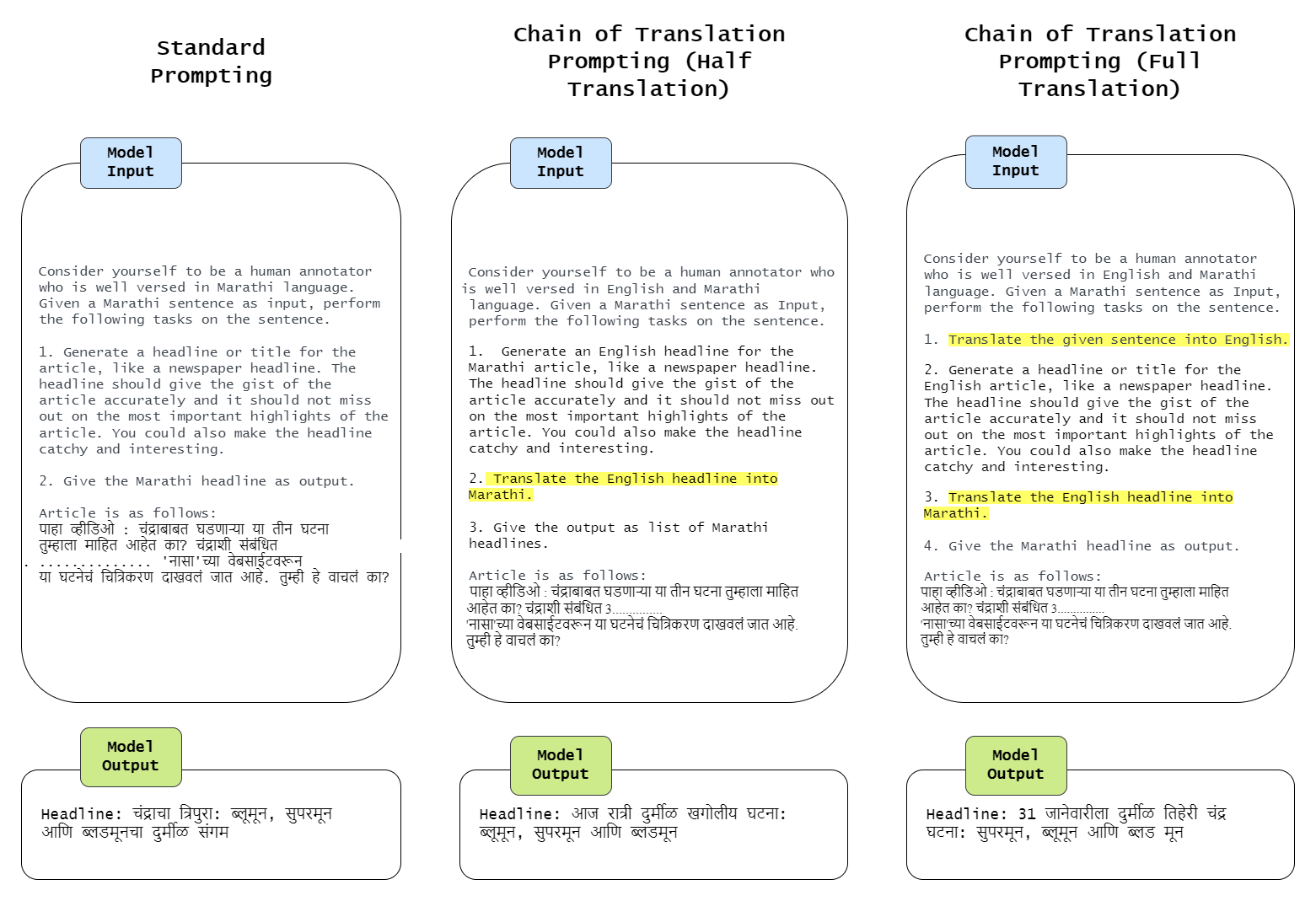}
  \caption{Prompt for Generation Task}
  \label{fig:ict_paper_fig}
\end{figure*}

\section {Related Work}
Natural language processing has improved significantly with the creation of sophisticated models like GPT-4, Llama3, and others. Nonetheless, insufficient representation and scarce data availability in pre-trained models continue to pose problems for low-resource languages\cite{b32}. Language diversity and data scarcity in low-resource contexts have shown to be challenges for traditional NLP techniques, which has prompted a quest for novel approaches that can make better use of already-existing data. \cite{b31} research highlighted the significance of creating NLP tools that are especially suited for low-resource languages while taking linguistic and cultural quirks into account.
\newline
A crucial component of developing NLP models for low-resource languages is dataset curation. In addition to collecting data, curators of datasets such as MahaSent, MahaHate, MahaNews-SHC, and CSEBUETNLP XLSum make sure that the data accurately reflects the linguistic diversity and cultural context of the language. Projects like \cite{b47} have brought attention to how crucial it is to provide high-quality datasets that accurately represent language use in everyday situations.
\newline
In multilingual natural language processing, cross-lingual transfer methods have demonstrated potential, especially when applied to low-resource language tasks. According to research like that of \cite{b34}, the concept of sharing parameters across languages allows models to acquire representations that function well in a variety of languages. This idea is important because it enables language models to use their English language skills to complete tasks in Marathi through the use of translation-based prompting, which is a type of cross-lingual transfer. Cross-lingual skills are supported by recent advances in multilingual models, such as mBERT \cite{b35} and XLM-R \cite{b36}, which lay a strong platform for further gains in low-resource language processing.
\newline
Prompting strategies have become an effective way to train large language models (LLMs) for particular tasks without requiring a lot of fine-tuning. According to \cite{b39}, well-crafted prompts can direct models such as GPT-3 to carry out a range of NLP tasks effectively. More research has been done on the subject of quick engineering's potential to induce desired behaviors in LLMs even in situations with limited resources by \cite{b38}. These methods have shown to be useful, particularly for languages and activities for which there is little to no direct training data.
\newline
Prompting is being used more and more for tasks like sentiment analysis and hate speech detection, which are essential for keeping an eye on public conversation and guaranteeing secure online spaces. Research on Pattern-Exploiting Training (PET) for such tasks was first presented by \cite{b40}, who showed how prompts could direct models to make context-based, nuanced predictions. This method is consistent with the findings of \cite{b41}, who also highlighted the benefit of model prompting for text categorization tasks in a variety of languages and domains.

\section{Methodology}

\subsection{Chain of Translation Prompting}

Our study introduces a novel approach called "Chain of Translation Prompting" aimed at enhancing the processing of Marathi, a low-resource language, using advanced language models like GPT-4o, GPT-4o Mini, Llama3-8B, Llama3-405B, and Gemma-9B. Recognizing the strong translation capabilities of these models, we leverage their ability to translate Marathi into English for improved processing. Directly prompting language models in Marathi has posed several challenges, primarily due to the scarcity of quality training data and the models' limitations in comprehending underrepresented languages. These challenges often result in sub-optimal performance on tasks such as sentiment analysis, hate speech classification, news categorization, and headline generation. Below, we outline the step-by-step methodology employed in our approach.

\begin{figure}[]
  \centering
\includegraphics
[width=\linewidth]
{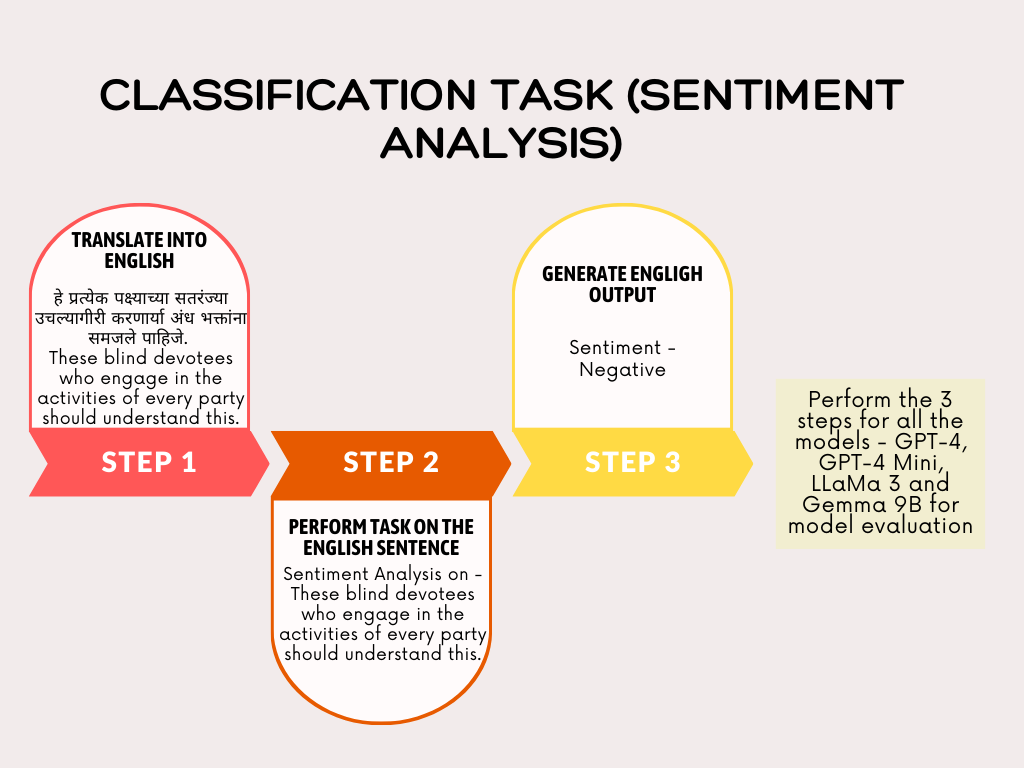}
\caption{Classification Task using Chain of Translation Prompting}
\label{fig:overview}
\end{figure}

\begin{enumerate}
    \item \textbf{Data Collection and Preparation:}
    We used datasets specific to Marathi language tasks, including MahaSent for sentiment analysis, MahaHate for hate speech classification, and MahaNews-SHC for subject categorization. For generative tasks, we used the CSEBUETNLP XLSum dataset to generate headlines.

    \item \textbf{Chain of Translation Prompting (CoTR) Technique:}
    Our methodology adapts a conventional translation approach used in developing low-resource NLP systems but applies it within the framework of large language model (LLM) prompts. Specifically, our method involves prompting the LLM to first translate the input text from Marathi into English, and then to execute the desired task on the translated English text.

    \item \textbf{Task Execution:}
    \begin{itemize}
        \item \textbf{Sentiment Analysis, Hate Speech Classification, and Subject Categorization}: For these classification tasks, the models categorize each sentence into predefined classes based on the task's requirements.
        \item \textbf{Generative Task}: We used GPT-4o, GPT-4o Mini, and Llama3-405b for the headline generation task. The three prompting strategies used for generating headlines are described below.
        \begin{enumerate}
            \item Without Translation: In this approach, headlines were generated directly from the original Marathi articles without any translation. This method aimed to assess the model’s capability to generate concise and impactful headlines in the source language.
            \item Full Translation: Here, the entire Marathi article was first translated into English. Headlines were then generated based on the translated English text. The generated English headlines were subsequently translated back into Marathi to evaluate their fidelity and relevance.
            \item Half Translation: Given the length and complexity of the articles, the half-translation method was employed to streamline the process. In this approach, English headlines were generated based on the Marathi articles without full translation. These English headlines were then translated back into Marathi. This method aimed to balance efficiency and accuracy by avoiding the need for extensive translation of the entire article.
        \end{enumerate}

    \end{itemize}
    \item \textbf{Direct Prompting}:
    To evaluate the effectiveness of the Chain of Translation Prompting, we compare its results against the traditional method of directly prompting the models to process the Marathi text without performing translation.
    \item \textbf{Google Translate + Prompting}:
    In this approach, Marathi sentences were translated into English using Google Translate. The translated English sentences, along with English prompts, were then used by LLMs to perform the desired classification tasks. This method represents a straightforward "translate-and-test" approach, serving as a baseline for comparison.

    \item \textbf{Evaluation Metrics}:
    The performance of the models is measured using conventional metrics, such as the ROUGE-L score for generative tasks. The ROUGE-L score assesses the quality of the generated text, like summaries or headlines, by calculating the overlap with reference text. It evaluates precision and recall by calculating the longest common subsequence (LCS) between the reference text and the generated output. ROUGE-L focuses on capturing the longest word sequences found in both texts, providing insights into the preservation of critical information and coherence.

For classification tasks, the model outputs are compared with ground truths, and the error percentage is reported.
\end{enumerate}

\begin{figure}[]
  \centering
\includegraphics
[width=\linewidth]
{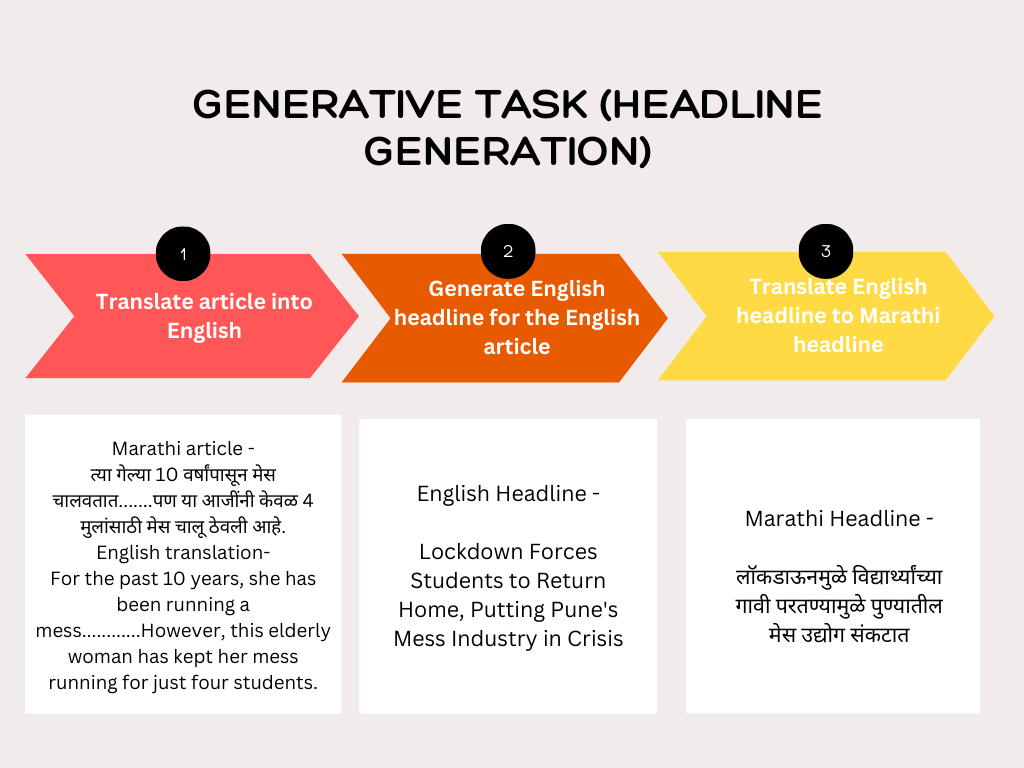}
\caption{Generative Task using Chain of Translation Prompting}
\label{fig:overview}
\end{figure}

 \begin{table*}[h]
\centering
\caption{Results on the MahaNews, MahaHate Dataset}
  \includegraphics[width=1\textwidth]{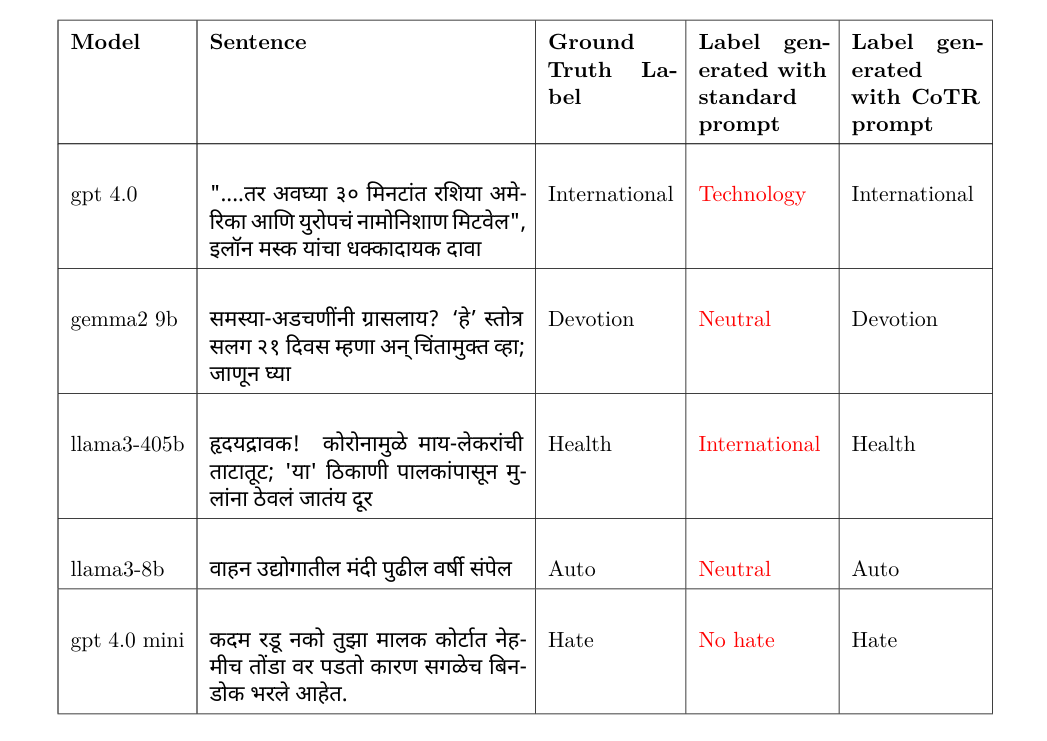}
  \label{tab:classification_examples}
\end{table*}

\subsection {Datasets Used}

\begin{enumerate}
    \item MahaSent-GT\footnote{\url{https://github.com/l3cube-pune/MarathiNLP/tree/main/L3Cube-MahaSent-MD}}:
    \newline
    We used a subset of the L3Cube-MahaSent-MD dataset \cite{b42}, which contains 14,000 annotated Marathi tweets. Three sentiment labels Positive, Negative, and Neutral are present in the dataset. In particular, we employed the MahaSent-GT portion of this dataset for sentiment analysis.
    \item MahaHate\footnote{\url{https://github.com/l3cube-pune/MarathiNLP/tree/main/L3Cube-MahaHate}}:
    \newline
    We used the L3Cube-MahaHate collection's MahaHate 2-Class dataset for our classification task \cite{b43}. It contains around 37500 annotated Marathi sentences. This dataset is divided into two categories: hate and non-hate. We employed the MahaHate 2-Class set for our classification task.
    \item MahaNews-SHC\footnote{\url{https://github.com/l3cube-pune/MarathiNLP/tree/main/L3Cube-MahaNews-SHC}}:
    \newline
    We analyzed Marathi news articles using the L3Cube-MahaNews-SHC dataset \cite{b44}. This dataset contains approximately 54,000 news articles spanning a wide range of topics and was used for the news classification task.
    \item XLSum\footnote{\url{https://huggingface.co/datasets/csebuetnlp/xlsum}}:
    \newline
    We focused on Marathi text headline creation for our study using the CSEBUETNLP XLSum dataset. The dataset offers a wide range of news stories linked with their associated headlines. Our objective was to enhance the accuracy of automated headline creation for Marathi news articles by utilizing this dataset.
\end{enumerate}

\subsection{Evaluation Methodology}
We performed our classification task on GPT-4o, GPT-4o Mini, Llama3-8B, Llama3-405B, and Gemma-9B.
\begin{enumerate}
    \item \textbf{GPT-4o:}
    \newline
     GPT-4o is developed by OpenAI, with ~1.8 trillion parameters (unofficial). It is a closed-source model and accessible through APIs provided by OpenAI. GPT-4o builds on the advancements of its previous versions, offering enhanced capabilities in natural language understanding, generation, and reasoning across a wide range of tasks.
    \item \textbf{GPT-4o Mini:}
    \newline
    GPT-4o Mini is a smaller, more lightweight version of GPT-4o. This model is closed-source. GPT-4o Mini is engineered to balance computational efficiency with performance, making it suitable for applications requiring faster inference times and lower resource consumption while maintaining a high level of language understanding.
    \item \textbf{Llama 3.1 8B / 405B:}
    \newline
    Llama 3.1 (Large Language Model for Multilingual Applications) is the third iteration in the Meta Llama series, designed with multiple variants, including a 405 billion parameter version and an 8 billion parameter version. These models are typically open-source. Llama3 models are optimized for multilingual tasks, incorporating vast and diverse datasets to improve performance across different languages.
    \item \textbf{Gemma-2 9B:}
    \newline
    Gemma-2 9B is an open-source language model with 9 billion parameters from Google. It strikes a balance between model size and performance, offering robust capabilities for both academic and practical applications.
\end{enumerate}

\begin{table}[h!]
\centering
\begin{tabularx}{\linewidth}{|X|X|X|X|}
\hline
Model & Without Translation & Half Translation & Full Translation \\ \hline
GPT-4o & 33.3 & 44 & 49 \\ \hline
GPT-4o mini & 21.34 & 21.72 & 22.22 \\ \hline
llama3-405b &  20.27 & 20.34 & 21.13 \\ \hline
\end{tabularx}
\caption{Rouge-L score in percentage for 3 approaches on the headline generation task on  CSEBUETNLP XLSum Dataset}
\label{tab:RougeLscore_generative}
\end{table}

\begin{table*}[h]
\centering
\caption{Error percentage in the classification task across 5 models (these are the weighted averages and the numbers are percentages). Standard Prompt - Prompt the LLM to perform the task using the given Marathi context. CoTR Prompt - Prompt the LLM to translate the Marathi context to English and then perform the task. Google Translate - Translate the Marathi context to English using Google Translate and then prompt the LLM to perform the task in English. }
\label{tab:classification_result}
\begin{tabular}{|p{1.78cm}|c|p{1.5cm}|p{1.3cm}|p{1.3cm}|p{1.3cm}|p{1.7cm}| p{1.8cm}|}
\hline
Model & Dataset & Standard Prompt & CoTR Prompt & Google Translate & Average Standard Prompt & Average CoTR Prompt & Avg Google Translate Prompt \\ \hline
\multirow{3}{*}{\centering\arraybackslash gpt-4o} & MahaSent & 20.38 & 18.44 & 25.00 & \multirow{3}{*}{\centering\arraybackslash 13.57} & \multirow{3}{*}{\centering\arraybackslash 11.25} & \multirow{3}{*}{\centering\arraybackslash 18.70}  \\ \cline{2-5}
                      & MahaNews & 3.06 & 2.04& 6.12 &  & &   \\ \cline{2-5}
                      & MahaHate & 16.83 & 12.87& 26.70 &  & &  \\ \hline
                      
\multirow{3}{*}{\centering\arraybackslash gpt-4o mini} &  MahaSent & 20.38 & 19.41 & 33.00 & \multirow{3}{*}{\centering\arraybackslash 20.19} & \multirow{3}{*}{\centering\arraybackslash 15.23}  & \multirow{3}{*}{\centering\arraybackslash 24.40} \\ \cline{2-5}
                      & MahaNews & 6.12 & 4.08 & 9.18 & & &  \\ \cline{2-5}
                      & MahaHate & 33.66 & 21.78 & 30.70 & & &  \\ \hline

\multirow{3}{*}{\centering\arraybackslash llama3-405b} & MahaSent & 31.06 & 27.18 & 22.00 & \multirow{3}{*}{\centering\arraybackslash 19.86} & \multirow{3}{*}{\centering\arraybackslash 16.22}  & \multirow{3}{*}{\centering\arraybackslash 18.70} \\ \cline{2-5}
                      & MahaNews & 7.14 & 6.12 & 6.12 & & &  \\ \cline{2-5}
                      & MahaHate & 20.89 & 14.85 & 27.72 & & &  \\ \hline

\multirow{3}{*}{\centering\arraybackslash llama3-8b} &  MahaSent & 35.92 & 27.18 & 30.00 & \multirow{3}{*}{\centering\arraybackslash 29.13} & \multirow{3}{*}{\centering\arraybackslash 23.84}  & \multirow{3}{*}{\centering\arraybackslash 24.00} \\ \cline{2-5}
                      & MahaNews & 10.20 & 7.14 & 9.18 & & & \\ \cline{2-5}
                      & MahaHate & 40.59 & 36.63 & 32.60 & & & \\ \hline

\multirow{3}{*}{\centering\arraybackslash gemma9b} &  MahaSent & 33.98 & 27.18 & 29.00 & \multirow{3}{*}{\centering\arraybackslash 22.18} & \multirow{3}{*}{\centering\arraybackslash 22.51}  & \multirow{3}{*}{\centering\arraybackslash 22.40} \\ \cline{2-5}
                      & MahaNews & 10.20 & 10.20 & 11.20 & & &  \\ \cline{2-5}
                      & MahaHate & 21.78 & 29.70 & 26.70 & & & \\ \hline

\end{tabular}
\end{table*}

\section{Results and Discussion}
Table~\ref{tab:RougeLscore_generative} and Table~\ref{tab:classification_result} show the analysis done on Standard Prompting and Chain of Translation Prompting.

\subsection{Classification Task} 
Approximately 100 sentences were selected from MahaSent-GT, MahaNews-SHC, and MahaHate. The large language models categorize each of the sentences into a predefined category. These results were compared with the ground truth values to calculate the error rate.
The error rate was calculated with the direct prompting approach and Chain of Translation prompting approach
The results are shown in Table~\ref{tab:classification_result}.
\newline
In the CoTR prompting approach, the error rate has reduced by 2.32\% in the GPT-4o model, by 3.64\% llama3-405b, by 5.29\% in llama3-8b and by 4.96\% in GPT-4o Mini. The error rate is slightly increased by 0.33\% in the Gemma-9B model.
\newline 
The error rate has been reduced by almost 5\% in llama3-8b and gpt4 mini models. Specifically, the CoTR prompting approach has significantly improved hate speech identification across all models except for Gemma-9B. In the hate speech classification task, Gemma-9B often failed to correctly translate hateful comments and, in some cases, omitted those parts entirely. However, compared to standard prompting, the number of misclassifications for the "Non-hate" class was lower when using CoTR.
\newline
The results from the CoTR approach show significant improvement over the standard Google Translate method as well. We manually reviewed the translated sentences and found out that translations by large language models (LLMs), such as GPT-4 and GPT-4 Mini, captured meanings and nuances more effectively than Google Translate. While LLMs conveyed the intended meaning with subtlety, Google Translate produced literal translations, which sometimes failed to capture the full sense of the sentences. 
\newline
For GPT-4 and GPT-4 Mini, the direct translation approach surpassed Google Translate’s performance, as the nuances of some of the sentences did not get extracted completely by the google translator. As GPT-4 and GPT-4 Mini are stronger models the direct prompting is working better than Google translator approach.
\newline
One sample detection with traditional prompting versus CoTR prompting from each of the four models has been attached in Table~\ref{tab:classification_examples}, where the output with CoTR prompting is the same as the ground truth.

 \begin{table*}[h]
\centering
\caption{Sample examples for the headline generation task using CoTR}
  \includegraphics[width=1\textwidth]{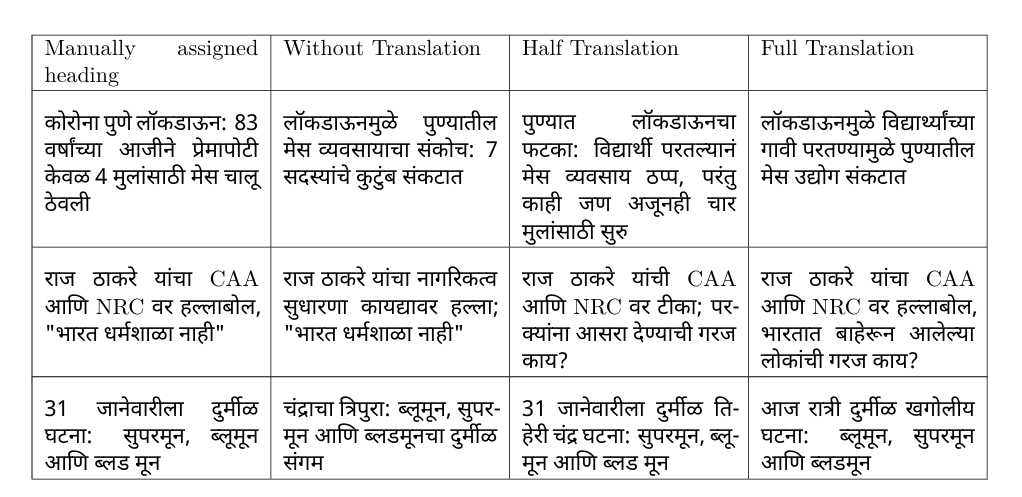}
  \label{tab:generative_examples}
\end{table*}

\subsection{Generation Task} 

The headlines from the Marathi news text were generated using traditional prompting and CoTR prompting (with half and full translation). The headlines were compared against the manually assigned headline and the Rouge-L score metric was used to calculate their similarity with the manually assigned headline.
The Rouge-L score for traditional prompting and CoTR prompting (half and full translation) are given in Table~\ref{tab:RougeLscore_generative}
\newline
We observed that GPT-4o delivered the best performance among all the models. GPT-4o Mini struggled to identify fine details in the articles, while Llama3-405B occasionally failed to provide the results in the specified format and produced some inaccurate translations. Overall, GPT-4o Mini and Llama-405B yielded similar outcomes.
\newline 
In general, we observe the following performance ranking for Marathi tasks: GPT-4o > GPT-4o Mini > Llama 3.1 405B > Gemma 2 9B > Llama 3.1 8B. The CoTR approach proves especially useful with smaller models and for complex tasks like hate speech detection and sentiment analysis.

\section {Future Work and Conclusion}
In summary, our study demonstrates that various prompting strategies, particularly the Chain of Translation (CoTR) method, effectively enhance Marathi language processing tasks. By applying these techniques to various classification and generation tasks, we have expanded the potential for more reliable and accurate NLP applications in Marathi. While CoTR improves model performance, it does so at the cost of generating more tokens.
\newline
In the future, we aim to enhance performance on Marathi language tasks by combining Chain of Thought (CoT) and Chain of Translation (CoTR) prompting strategies. Our goal is to achieve context-aware and precise responses for complex tasks like sentiment analysis, hate speech detection, and subject classification. CoT allows models to break down complex tasks into simpler steps, while CoTR leverages translation from Marathi to English, where more accurate models can be employed. Together, these strategies should create a robust framework that improves model performance and reliability in Marathi NLP tasks.
\newline
This approach can further be used for other low-resource Indic languages.

\section*{Acknowledgments}

This work was done under the mentorship of Mr. Raviraj Joshi (Mentor, L3Cube Pune). We would like to express our gratitude towards him for his continuous support and encouragement.

\bibliography{main}
\nocite{*}

\end{document}